\pdfoutput=1

\documentclass[11pt]{article}

\usepackage[preprint]{acl}

\usepackage{times}
\usepackage{latexsym}

\usepackage[T1]{fontenc}
\usepackage{amsmath}
\usepackage{orcidlink}
\usepackage{amssymb}
\usepackage{amsfonts}
\usepackage{bm}
\usepackage{booktabs}
\usepackage{subfig}

\usepackage[utf8]{inputenc}

\usepackage{microtype}

\usepackage{inconsolata}

\usepackage{graphicx}

\title{Affinity and Diversity: A Unified Metric for Demonstration Selection via Internal Representations}

\author{Mariko Kato\orcidlink{0009-0000-3184-0334}${}^{1}$\phantom{111}
Hakaze Cho\orcidlink{0000-0002-7127-1954}${}^{1}$\phantom{111}
Yoshihiro Sakai${}^{1}$\phantom{111}
\textbf{Naoya Inoue}${}^{1,2}$\\
${}^{1}$Japan Advanced Institute of Science and Technology\phantom{111}${}^{2}$RIKEN\\
\texttt{mariko.k@jaist.ac.jp}}

\begin{document}
\maketitle

\begin{abstract}
The performance of \textbf{I}n-\textbf{c}ontext \textbf{L}earning (ICL) is highly sensitive to the selected demonstrations. Existing approaches to demonstration selection optimize different objectives, yielding inconsistent results. To address this, we propose a unified metric--affinity and diversity--that leverages ICL model's internal representations. Our experiments show that both affinity and diversity strongly correlate with test accuracies, indicating their effectiveness for demonstration selection. Moreover, we show that our proposed metrics align well with various previous works to unify the inconsistency.
\end{abstract}

\section{Introduction}

\textbf{L}anguage \textbf{M}odels (LMs) show \textbf{I}n-\textbf{C}ontext \textbf{L}earning (ICL) ability~\cite{dong2024surveyincontextlearning}, learning to solve tasks without updating model parameters by processing a query along with demonstrations of input-label pairs.
The performance of ICL is highly sensitive to the quality of demonstrations~\cite{liu2021makesgoodincontextexamples}, and previous work has proposed several strategies for selecting better demonstrations.

One prominent approach is to select demonstrations based on their similarity to queries.
Here, the similarity is computed by models \emph{independent of the ICL-executed LMs}, e.g., off-the-shelf document retrievers~\cite{rubin2022learningretrievepromptsincontext}, such as BM25~\cite{bm25}, fine-tuned document retrievers~\cite{luo2024incontextlearningretrieveddemonstrations}, and encoder-based pretrained LMs~\cite{chen2024bgem3embeddingmultilingualmultifunctionality}.
While these previous methods have improved ICL performance, we find that they capture different aspects of demonstration quality and do not converge on a consensus measure (Fig.~\ref{fig:analysis}, Left).
Developing a unified metric that integrates various metrics leads to a deeper understanding of demonstration quality and further enhances ICL performance.

\begin{figure}[t]
    \centering
    \includegraphics[width=1\linewidth]{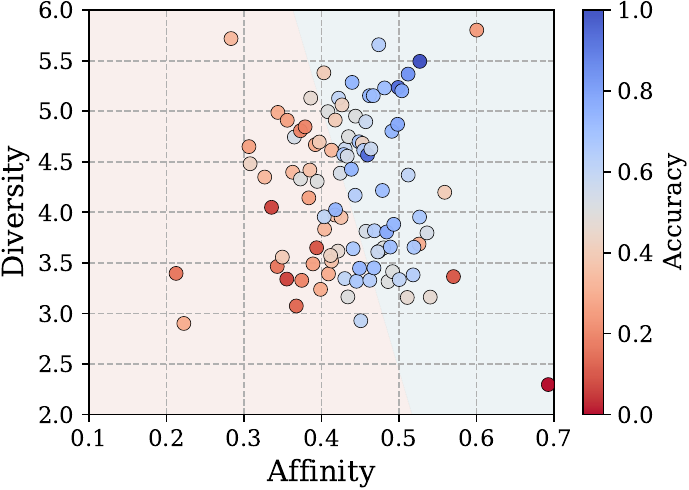}
    \caption{The Affinity and Diversity of the demonstrations in TREC, SST5, TEE on $k=16$. The colors of the circles refer to the accuracy of the classification tasks. The line and background color refer to the decision boundary to predict labels by affinity and diversity. The larger the affinity and diversity, the higher the accuracy tends to be.  }
    \label{fig:Llama3-8b_Affinity-Diversity_scatter}
\end{figure}

Therefore, in this paper, we propose a novel approach that \emph{leverages the ICL model’s internal representations} to unify previous selection methods.
We first identify a self-attention head that is critical for ICL, and on the subspace defined by the $W_Q^\top W_K$ of this attention head, we define two new metrics: (i) \emph{affinity} between a query and demonstrations and (ii) \emph{diversity} among demonstrations.
Our experiments show that proposed metrics correlate with existing demonstration selection methods (Fig.\ref{fig:analysis}, Left) and are useful for identifying better demonstrations (Fig.~\ref{fig:Llama3-8b_Affinity-Diversity_scatter}).

\vspace{0.5\baselineskip}
\noindent\textbf{Our contributions are:}




\begin{itemize}
    \item We propose internal representation-based affinity and diversity as a better joint metric on ICL for demonstration selection (\S\ref{sec:mainresults}), which unifies the previous selection methods (\S\ref{sec:correlation_between_aff/div_previous}).
    \item We empirically show that previous demonstration selection methods focus on different aspects of selected demonstrations, showing that they are not always positively correlated with other selection methods (\S\ref{sec:correlation_among_previous}).
\end{itemize}

\begin{figure}[t]
    \centering
    \includegraphics[width=1\linewidth]{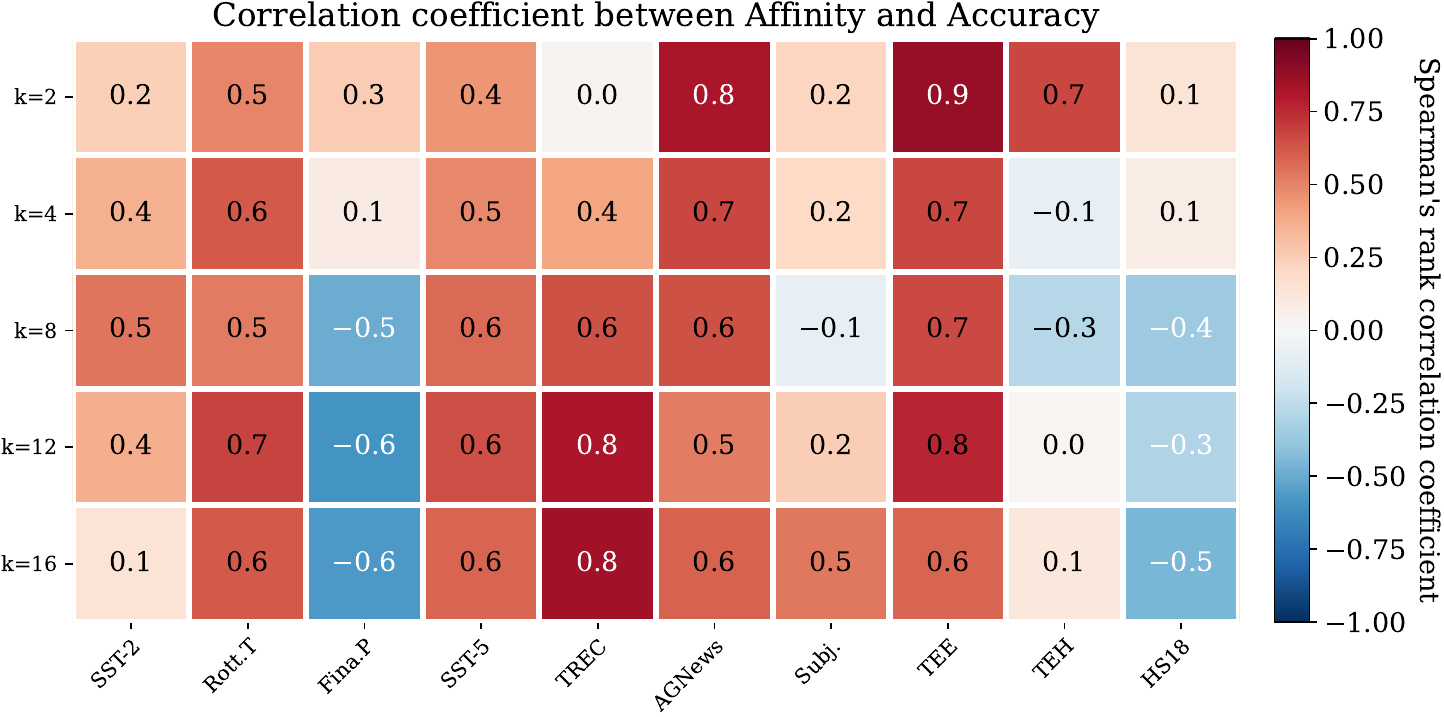}
    \caption{The correlation coefficient between affinity and accuracy}
    \label{fig:Llama3-8b_affinity_accuracy_heatmap}
\end{figure}

\section{Background}
\subsection{In-context Learning}
Given $k$ input-label pairs (\textit{demonstrations}) and a \textit{query} for a classification task, the demonstrations and query are concatenated in natural language form and fed to LMs (e.g., for $k=2$, ``Good movies. Label: Positive. That's too cruel. Label: Negative. I like it. Label: ''). Here, ``: '' serves as a \textit{forerunner token} to concatenate inputs and labels, and trigger the prediction of the label tokens.
The LMs return a probability distribution over the next tokens, and ICL selects the token with the highest probability as the final prediction. 

\subsection{Induction Circuit}  
An induction circuit is an abstraction of some attention heads to lead the inference of ICL~\cite{elhage2021mathematical}, which consists of several interacting attention heads across different layers: (i) \textit{previous token heads}, which copy information from previous tokens to the current token, and (ii) \textit{induction heads}, which attend to tokens based on context and boost the probability of predicting token $[B]$ when $[A][B]...[A']$ is provided as input. In this paper, we find the most effective induction head, and define the affinity-diversity metrics on the $W_Q^\top W_K$ mappings of this head.

\subsection{Demonstration Selection Methods}
There are two main approaches for retrieving demonstrations. One is to use off-the-shelf retrievers, such as BM25~\cite{bm25} or BGE M3~\cite{chen2024bgem3embeddingmultilingualmultifunctionality}. Off-the-shelf retrievers approaches may be sub-optimal since they are not finetuned for specific tasks.
Another approach is to train retrievers, e.g., using encoder-based LMs, based on supervision signals from ICL models. To optimize such retrievers, various loss functions (e.g., List-wise Ranking Loss~\cite{li2023unifieddemonstrationretrieverincontext} and InfoNCE Loss~\cite{rubin2022learningretrievepromptsincontext}) and training strategies (e.g., iterative training or contrastive training) are employed. Note that while learning-based methods learn signals from ICL models during training, they solely rely on the trained retriever during ICL. However, in \S\ref{sec:correlation_among_previous}, we show that there is no consistent correlation between these previous approaches, leading to disagreement in the selected demonstrations across different objectives and optimization methods, that should be unified for consistent demonstration selection.
\begin{figure}[t]
    \centering
    \includegraphics[width=0.97\linewidth]{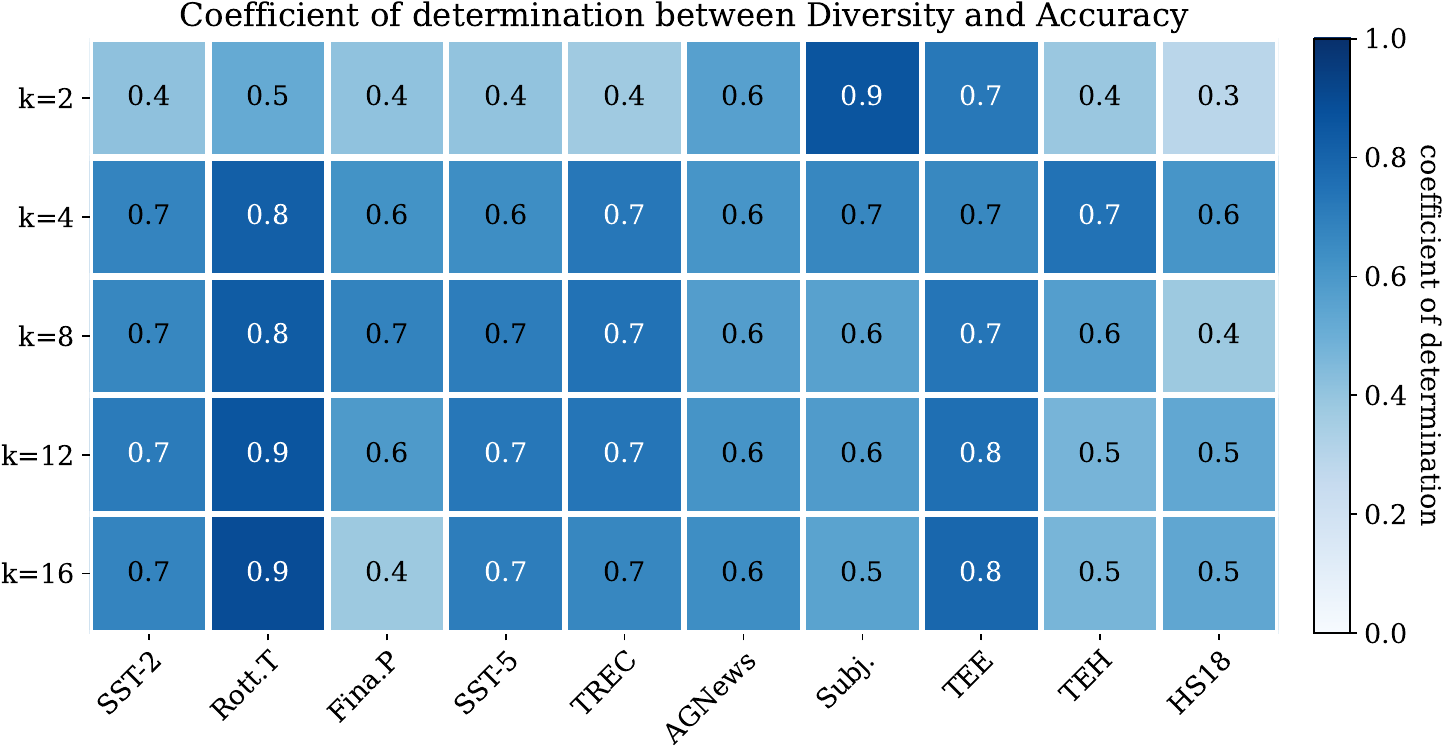}
    \caption{The coefficient of determination between diversity and accuracy}
    \label{fig:Llama3-8b_diversity_accyract_heatmap}
\end{figure}

\begin{figure*}
    \centering
    \begin{minipage}[b]{0.68\linewidth}
    \centering
        \includegraphics[width=1\linewidth]{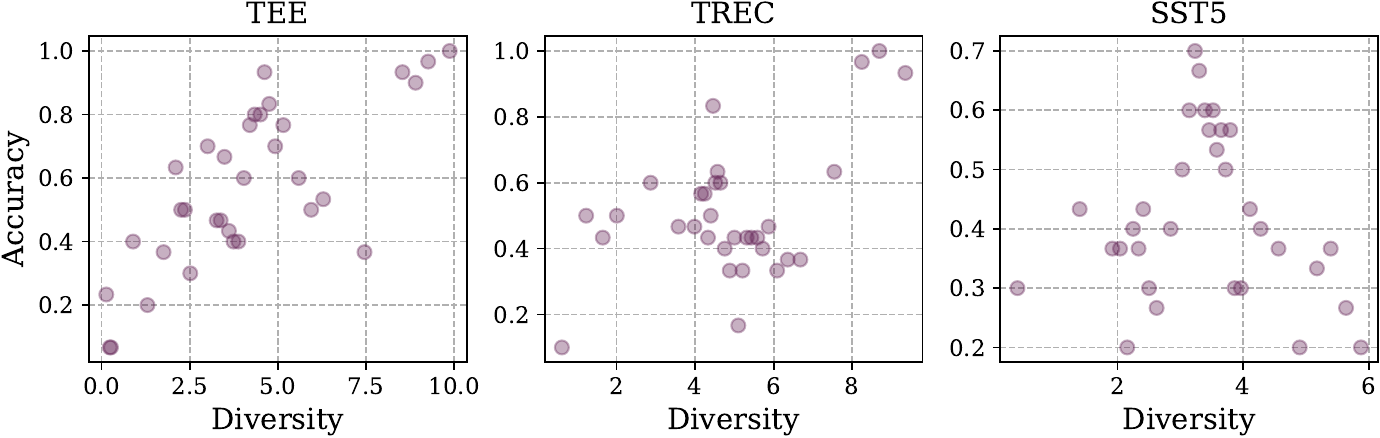}
    \end{minipage}
    \hspace{0.05\linewidth}
    \begin{minipage}[b]{0.255\linewidth}
    \centering
        \includegraphics[width=1\linewidth]{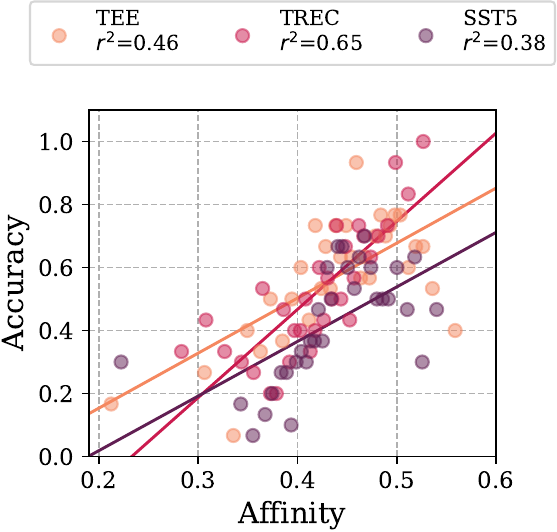}
    \end{minipage}
    
    \caption{\textbf{Left}: The tendency of diversity to accuracy on $k=16$. \textbf{Right}: The tendency of affinity to accuracy on $k=16$. }
    \label{fig:sec4-scatter}
\end{figure*}

\section{Proposed Metrics: Affinity, Diversity}

Since induction circuits play a crucial role in ICL,
we hypothesize that induction circuits can also be used to assess the quality of demonstrations.
We first identify induction heads (\S\ref{sec:step1}) and then compute affinity and diversity in their subspace (\S\ref{sec:step2}).

\subsection{Step 1: Extract Internal Representation}
\label{sec:step1}

To identify induction heads, we follow \citet{cho2024revisitingincontextlearninginference}:
for each attention head $h$ at layer $l$, we compute $s(h)$, the sum of attention scores from the last token of the query to all the correct label tokens (i.e., tokens that match the ground-truth label of the query) in the demonstrations, and identify ``the best induction head'' as the head $\hat{h}$ at layer $\hat{l}$ with the highest $s(\hat{h})$.

We then extract the label token representations $\left\{\bm{d}_\mathrm{label}^{(i)}\right\}_{i=1}^{k}$ of each demonstration $i$ and the last token's representation $\bm{d}_\mathrm{q}$ of the query from the best induction head $\hat{h}$. In detail, given a token index $j$ and the hidden state $\bm{h}_j^{\hat{l}}$ of $j$-th token from the previous layer of $\hat{h}$ after the layer normalization, we extract the inner representation of $j$-th token as follows:
\begin{equation}
    \bm{d}_j = W_Q^{\hat{h},\top} W_K^{\hat{h}} \bm{h}_j^{\hat{l}},
\end{equation}
where $W_Q^{\hat{h}}$ and $W_K^{\hat{h}}$ are the query projection and key projection of attention head $\hat{h}$.

\subsection{Step 2: Compute Affinity and Diversity}
\label{sec:step2}

\subsubsection{Affinity}
We define affinity as the mean of the cosine similarity between all the label token representations and the query representation as follows:
\begin{equation}\label{eq:aff}
    \mathrm{Aff}\left[\bm{d}_\mathrm{q}, \left\{\bm{d}_{\mathrm{label}}^{(i)}\right\}_{i=1}^k\right] = \frac{1}{k}\sum_{i=1}^k \mathrm{cos}\left[\bm{d}_\mathrm{q},\bm{d}_\mathrm{label}^{(i)}\right]
\end{equation}

\subsubsection{Diversity}
We define diversity as the variance (the trace of the covariance matrix) across the label token representations of all demonstrations as follows:
\begin{equation}\label{eq:cov}
    \mathrm{Div}\left[\left\{\bm{d}_{\mathrm{label}}^{(i)}\right\}_{i=1}^k\right] = \frac{1}{k}\mathrm{tr}\left[\mathop{\mathbb{D}}\limits_{i\in[1, k]}\left[\bm{d}_{\mathrm{label}}^{(i)}\right]\right]
\end{equation}
Here, $\mathbb{D}$ represents the covariance operator.

\begin{figure*}
    \centering
    \begin{minipage}[b]{0.29\linewidth}
        \centering
        \includegraphics[width=\linewidth]{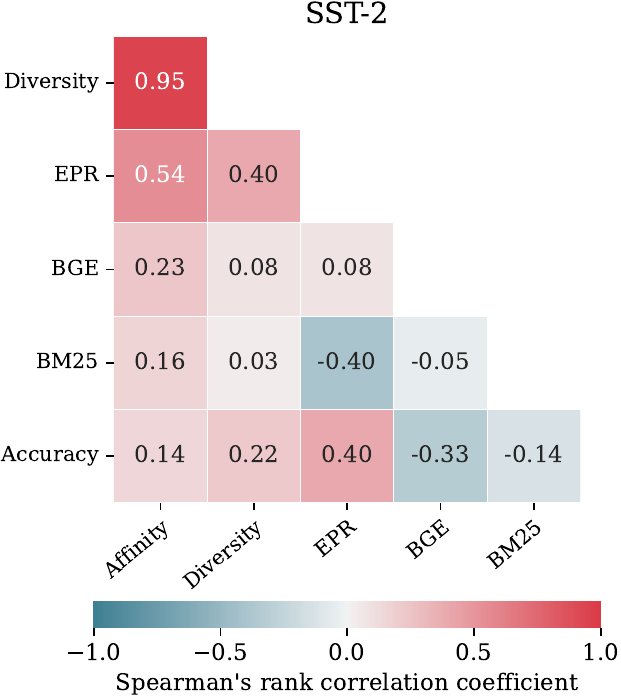}

    \end{minipage}
    \hspace{0.05\linewidth}
    \begin{minipage}[b]{0.6\linewidth}
        \centering
        \includegraphics[width=\linewidth]{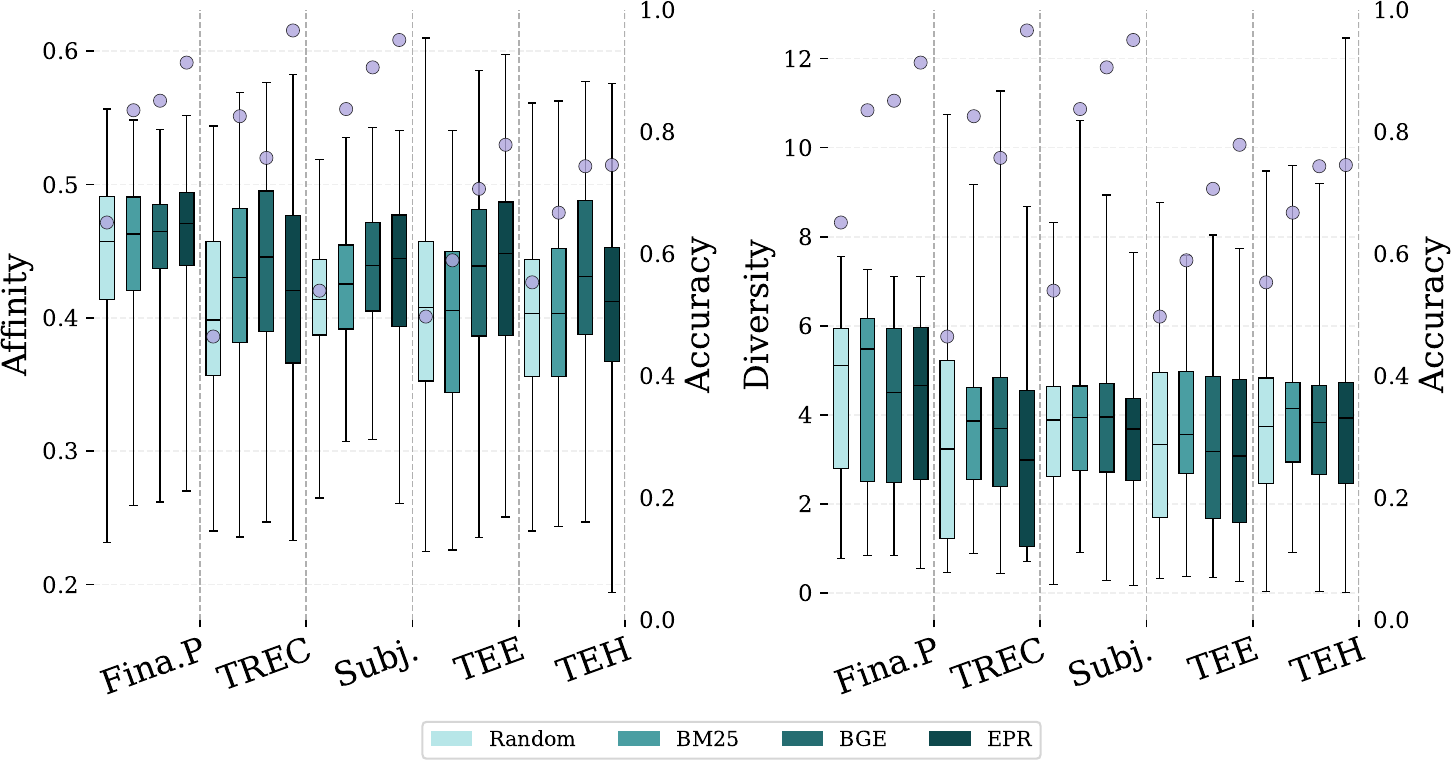}

    \end{minipage}

    \vspace{-0.3\baselineskip}
    \caption{\textbf{Left}: The Spearman's rank correlation coefficient of the similarity scores, affinity, diversity, and accuracy of $k=16$ on SST2. \textbf{Middle}: The affinity of selected demonstrations by each selection method on $k=2$. \textbf{Right}: The diversity of selected demonstrations by each selection method on $k=2$.}
    \label{fig:analysis}
    \vspace{-0.8\baselineskip}
\end{figure*}

\section{Experiments}

We demonstrate that affinity and diversity serve as effective metrics for demonstration selection.

\subsection{Experimental Settings}
\label{sec:settings}

\paragraph{Model.} We conduct experiments on Llama 3 8B~\cite{llama3modelcard}. The model parameters are loaded from \verb|HuggingFace|.

\paragraph{Dataset.} For all experiments, we use 10 classification datasets. For details of the dataset, please refer to Appendix~\ref{appendix:dataset}. We use $k = 2, 4, 8, 12, 16$, and input sequences are built by library StaICC~\cite{cho2025staicc}.

\paragraph{Evaluation.} For each test instance, we randomly sample $k$ demonstrations, run ICL, and record the prediction. Next, we sort all instances based on their affinity or diversity values and group them into bins of 30 instances each. For each bin, we calculate the average affinity or diversity, and also the accuracy. These averages and accuracies are then used to compute the correlation between the proposed metrics and accuracy. For affinity, we use Spearman's rank correlation coefficient. For diversity, we apply Ridge regression with a Laplacian kernel to capture non-linear relationships, with the $R^2$ coefficient as the measure of goodness-of-fit.

\subsection{Main Results: Affinity and Diversity Measure the Effectiveness of Demonstrations}
\label{sec:mainresults}

The Spearman's rank correlation coefficient for affinity and $R^2$ coefficient for diversity are shown in Fig.~\ref{fig:Llama3-8b_affinity_accuracy_heatmap} and Fig.~\ref{fig:Llama3-8b_diversity_accyract_heatmap}.
These indicate that affinity shows a positive correlation across various tasks, and diversity achieves a high $R^2$ coefficient in nearly all tasks. Fig.~\ref{fig:sec4-scatter} (Left) and Fig.~\ref{fig:sec4-scatter} (Right) provide examples of diversity/affinity-accuracy scatter plots, which further support these trends.


\section{Analysis}


Next, we show that affinity and diversity strongly correlate with the scores from previous demonstration selection methods, addressing the inconsistency issue in the previous work (\S\ref{sec:correlation_between_aff/div_previous}).
We also show that the scores from previous demonstration selection methods disagree with each other (\S\ref{sec:correlation_among_previous}).  Moreover, the demonstrations selected by previous work practically improve affinity, but not diversity. These observations suggest that it is required a new demonstration selection method based on affinity and diversity (\S\ref{sec:previous_improve_aff}).

\subsection{Experimental Setup}
We use three previous methods to compare affinity and diversity, BM25 and BGE-M3 for training-free methods, and EPR for training methods. For details of the previous methods, please refer to Appendix~\ref{appendix:previous-methods}. Other settings are the same as \S\ref{sec:settings}.


\subsection{Affinity and Diversity Correlate with the Score of Previous Methods}
\label{sec:correlation_between_aff/div_previous}
We measure the similarity scores from the previous methods using the same prompts described in \S\ref{sec:settings} and compute the Spearman's rank correlation among these similarity scores, accuracy, affinity, and diversity. The results of $k=16$ on SST2 are shown in Fig.~\ref{fig:analysis} (Left), where both affinity and diversity show a positive correlation with the similarity scores and accuracy. This indicates that affinity and diversity consistently measure the effectiveness of the demonstrations in terms of accuracy. 

\subsection{Previous Methods are Not Always Positively Correlated with Each Other}
\label{sec:correlation_among_previous}
Meanwhile, no consistent positive correlation is observed among the similarity scores from previous selection methods. Even worse, in some cases, negative correlations (e.g., EPR and BM25) are observed, suggesting that they may not consistently produce optimal results.
EPR shows a positive correlation with BGE, likely due to their reliance on a BERT-based encoder.


\subsection{Better Selection of Demonstrations Improves Affinity and Diversity}
\label{sec:previous_improve_aff}
In this section, we evaluate the previous demonstration selection methods on the proposed affinity and diversity, and show that affinity and diversity are improved by the previous methods. We build prompts with the same query as \S\ref{sec:settings} select demonstrations by previous methods and input them into an LM to measure the accuracy, affinity, and diversity.

The results are shown in Fig.~\ref{fig:analysis} (Middle) for the affinity and accuracy, where better accuracy co-occurrence with greater affinity, while, when no improvement is observed in the affinity, then no accuracy can be observed in the accuracy, on some of the scenarios. Moreover, the results of diversity are shown in Fig.~\ref{fig:analysis} (Left), with a less significant co-occurrence between better accuracy and greater diversity. We infer that the reason is: existing methods select demonstrations based on their similarity to the query, without a focus on the diversity, showing a possibility towards better selection methods based on the joint metric of affinity and diversity. Due to space limitations and computational resources, we leave the demonstration selection method as future work.

\section{Conclusion}
In summary, we propose affinity and diversity to evaluate demonstration selections in the ICL scenario. Our experiments show that affinity and diversity consistently measure the effectiveness of the demonstration well, raising the possibility of better demonstration selection methods.

\section{Limitations}
Due to computability limitations, we are not able to compare the performance of affinity and diversity with the learning-based retriever for diversity or order of demonstrations.

\newpage
\bibliography{custom}

\begin{thebibliography}{21}
\providecommand{\natexlab}[1]{#1}

\bibitem[{AI@Meta(2024)}]{llama3modelcard}
AI@Meta. 2024.
\newblock \href {https://github.com/meta-llama/llama3/blob/main/MODEL_CARD.md} {Llama 3 model card}.

\bibitem[{Basile et~al.(2019)Basile, Bosco, Fersini, Nozza, Patti, Rangel~Pardo, Rosso, and Sanguinetti}]{teh}
Valerio Basile, Cristina Bosco, Elisabetta Fersini, Debora Nozza, Viviana Patti, Francisco~Manuel Rangel~Pardo, Paolo Rosso, and Manuela Sanguinetti. 2019.
\newblock \href {https://doi.org/10.18653/v1/S19-2007} {{S}em{E}val-2019 task 5: Multilingual detection of hate speech against immigrants and women in {T}witter}.
\newblock In \emph{Proceedings of the 13th International Workshop on Semantic Evaluation}, pages 54--63, Minneapolis, Minnesota, USA. Association for Computational Linguistics.

\bibitem[{Chen et~al.(2024)Chen, Xiao, Zhang, Luo, Lian, and Liu}]{chen2024bgem3embeddingmultilingualmultifunctionality}
Jianlv Chen, Shitao Xiao, Peitian Zhang, Kun Luo, Defu Lian, and Zheng Liu. 2024.
\newblock \href {https://arxiv.org/abs/2402.03216} {Bge m3-embedding: Multi-lingual, multi-functionality, multi-granularity text embeddings through self-knowledge distillation}.
\newblock \emph{Preprint}, arXiv:2402.03216.

\bibitem[{Cho and Inoue(2025)}]{cho2025staicc}
Hakaze Cho and Naoya Inoue. 2025.
\newblock Staicc: Standardized evaluation for classification task in in-context learning.
\newblock \emph{arXiv preprint arXiv:2501.15708}.

\bibitem[{Cho et~al.(2025)Cho, Kato, Sakai, and Inoue}]{cho2024revisitingincontextlearninginference}
Hakaze Cho, Mariko Kato, Yoshihiro Sakai, and Naoya Inoue. 2025.
\newblock \href {https://arxiv.org/abs/2410.04468} {Revisiting in-context learning inference circuit in large language models}.
\newblock In \emph{The Thirteenth International Conference on Learning Representations}.

\bibitem[{Conneau and Kiela(2018)}]{subj}
Alexis Conneau and Douwe Kiela. 2018.
\newblock Senteval: An evaluation toolkit for universal sentence representations.
\newblock \emph{arXiv preprint arXiv:1803.05449}.

\bibitem[{de~Gibert et~al.(2018)de~Gibert, Perez, Garc{\'\i}a-Pablos, and Cuadros}]{hs18}
Ona de~Gibert, Naiara Perez, Aitor Garc{\'\i}a-Pablos, and Montse Cuadros. 2018.
\newblock \href {https://doi.org/10.18653/v1/W18-5102} {{Hate Speech Dataset from a White Supremacy Forum}}.
\newblock In \emph{Proceedings of the 2nd Workshop on Abusive Language Online ({ALW}2)}, pages 11--20, Brussels, Belgium. Association for Computational Linguistics.

\bibitem[{Dong et~al.(2024)Dong, Li, Dai, Zheng, Ma, Li, Xia, Xu, Wu, Liu, Chang, Sun, Li, and Sui}]{dong2024surveyincontextlearning}
Qingxiu Dong, Lei Li, Damai Dai, Ce~Zheng, Jingyuan Ma, Rui Li, Heming Xia, Jingjing Xu, Zhiyong Wu, Tianyu Liu, Baobao Chang, Xu~Sun, Lei Li, and Zhifang Sui. 2024.
\newblock \href {https://arxiv.org/abs/2301.00234} {A survey on in-context learning}.
\newblock \emph{Preprint}, arXiv:2301.00234.

\bibitem[{Elhage et~al.(2021)Elhage, Nanda, Olsson, Henighan, Joseph, Mann, Askell, Bai, Chen, Conerly et~al.}]{elhage2021mathematical}
Nelson Elhage, Neel Nanda, Catherine Olsson, Tom Henighan, Nicholas Joseph, Ben Mann, Amanda Askell, Yuntao Bai, Anna Chen, Tom Conerly, et~al. 2021.
\newblock \href {https://transformer-circuits.pub/2021/framework/index.html} {A mathematical framework for transformer circuits}.
\newblock \emph{Transformer Circuits Thread}, 1(1):12.

\bibitem[{Hovy et~al.(2001)Hovy, Gerber, Hermjakob, Lin, and Ravichandran}]{trec2}
Eduard Hovy, Laurie Gerber, Ulf Hermjakob, Chin-Yew Lin, and Deepak Ravichandran. 2001.
\newblock \href {https://www.aclweb.org/anthology/H01-1069} {Toward semantics-based answer pinpointing}.
\newblock In \emph{Proceedings of the First International Conference on Human Language Technology Research}.

\bibitem[{Li et~al.(2023)Li, Lv, Yan, Lin, Zhu, Ni, Xie, Wang, and Qiu}]{li2023unifieddemonstrationretrieverincontext}
Xiaonan Li, Kai Lv, Hang Yan, Tianyang Lin, Wei Zhu, Yuan Ni, Guotong Xie, Xiaoling Wang, and Xipeng Qiu. 2023.
\newblock \href {https://arxiv.org/abs/2305.04320} {Unified demonstration retriever for in-context learning}.
\newblock \emph{Preprint}, arXiv:2305.04320.

\bibitem[{Li and Roth(2002)}]{trec1}
Xin Li and Dan Roth. 2002.
\newblock \href {https://www.aclweb.org/anthology/C02-1150} {Learning question classifiers}.
\newblock In \emph{{COLING} 2002: The 19th International Conference on Computational Linguistics}.

\bibitem[{Liu et~al.(2021)Liu, Shen, Zhang, Dolan, Carin, and Chen}]{liu2021makesgoodincontextexamples}
Jiachang Liu, Dinghan Shen, Yizhe Zhang, Bill Dolan, Lawrence Carin, and Weizhu Chen. 2021.
\newblock \href {https://arxiv.org/abs/2101.06804} {What makes good in-context examples for gpt-$3$?}
\newblock \emph{Preprint}, arXiv:2101.06804.

\bibitem[{Luo et~al.(2024)Luo, Xu, Liu, Pasupat, and Kazemi}]{luo2024incontextlearningretrieveddemonstrations}
Man Luo, Xin Xu, Yue Liu, Panupong Pasupat, and Mehran Kazemi. 2024.
\newblock \href {https://arxiv.org/abs/2401.11624} {In-context learning with retrieved demonstrations for language models: A survey}.
\newblock \emph{Preprint}, arXiv:2401.11624.

\bibitem[{Malo et~al.(2014)Malo, Sinha, Korhonen, Wallenius, and Takala}]{fp}
P.~Malo, A.~Sinha, P.~Korhonen, J.~Wallenius, and P.~Takala. 2014.
\newblock \href {https://arxiv.org/abs/1307.5336} {Good debt or bad debt: Detecting semantic orientations in economic texts}.
\newblock \emph{Journal of the Association for Information Science and Technology}, 65.

\bibitem[{Mohammad et~al.(2018)Mohammad, Bravo-Marquez, Salameh, and Kiritchenko}]{tee}
Saif Mohammad, Felipe Bravo-Marquez, Mohammad Salameh, and Svetlana Kiritchenko. 2018.
\newblock Semeval-2018 task 1: Affect in tweets.
\newblock In \emph{Proceedings of the 12th international workshop on semantic evaluation}, pages 1--17.

\bibitem[{Pang and Lee(2005)}]{rotten}
Bo~Pang and Lillian Lee. 2005.
\newblock Seeing stars: Exploiting class relationships for sentiment categorization with respect to rating scales.
\newblock In \emph{Proceedings of the ACL}.

\bibitem[{Robertson and Zaragoza(2009)}]{bm25}
Stephen Robertson and Hugo Zaragoza. 2009.
\newblock \href {https://doi.org/10.1561/1500000019} {The probabilistic relevance framework: Bm25 and beyond}.
\newblock \emph{Found. Trends Inf. Retr.}, 3(4):333–389.

\bibitem[{Rubin et~al.(2022)Rubin, Herzig, and Berant}]{rubin2022learningretrievepromptsincontext}
Ohad Rubin, Jonathan Herzig, and Jonathan Berant. 2022.
\newblock \href {https://arxiv.org/abs/2112.08633} {Learning to retrieve prompts for in-context learning}.
\newblock \emph{Preprint}, arXiv:2112.08633.

\bibitem[{Socher et~al.(2013)Socher, Perelygin, Wu, Chuang, Manning, Ng, and Potts}]{sst2}
Richard Socher, Alex Perelygin, Jean Wu, Jason Chuang, Christopher~D. Manning, Andrew Ng, and Christopher Potts. 2013.
\newblock \href {https://www.aclweb.org/anthology/D13-1170} {Recursive deep models for semantic compositionality over a sentiment treebank}.
\newblock In \emph{Proceedings of the 2013 Conference on Empirical Methods in Natural Language Processing}, pages 1631--1642, Seattle, Washington, USA. Association for Computational Linguistics.

\bibitem[{Zhang et~al.(2015)Zhang, Zhao, and LeCun}]{agnews}
Xiang Zhang, Junbo Zhao, and Yann LeCun. 2015.
\newblock \href {https://arxiv.org/abs/1509.01626} {Character-level convolutional networks for text classification}.
\newblock \emph{Advances in neural information processing systems}, 28.

\end{thebibliography}

\newpage
\appendix
\section{Experimental Details}
\subsection{Datasets}
\label{appendix:dataset}
We build ICL-formed test inputs from 10 classification tasks datasets: GLUE-SST2 (SST2)~\cite{sst2}, Rotten tomatoes (Rott.T)~\cite{rotten}, Finacial Phrasebank (Fina.P)~\cite{fp}, Stanford Sentiment Treebank (SST5)~\cite{sst2}, TREC (TREC)~\cite{trec1, trec2}, AGNews (AGNews)~\cite{agnews}, Subjective (Subjective)~\cite{subj}, Tweet Eval Emotion (TEE)~\cite{tee}, Tweet Eval Hate (TEH)~\cite{teh}, Hate Speech 18 (HS18)~\cite{hs18}.

\subsection{Previous methods}
\label{appendix:previous-methods}
We conduct experiments to compare affinity and diversity using previous methods:
\begin{itemize}
    \item BM25: selecting the demonstrations with the similarity score to query, by an expanded TF-IDF (BM25). 
    \item BGE M3: selecting the demonstrations with the most cosine similarity between the encoding vectors of the demonstrations and query, by BGE M3. The model parameters are loaded from \verb|HuggingFace|.
    \item \textbf{E}fficient \textbf{P}rompt \textbf{R}etrieval (EPR)~\cite{rubin2022learningretrievepromptsincontext}: selecting the same way as BGE M3, by the dense encoder trained to retrieve a better demonstration with each ICL datasets. 
\end{itemize}

\section{Other datasets experiment results}
\begin{figure*}[h]
    \centering
    \includegraphics[width=1\linewidth]{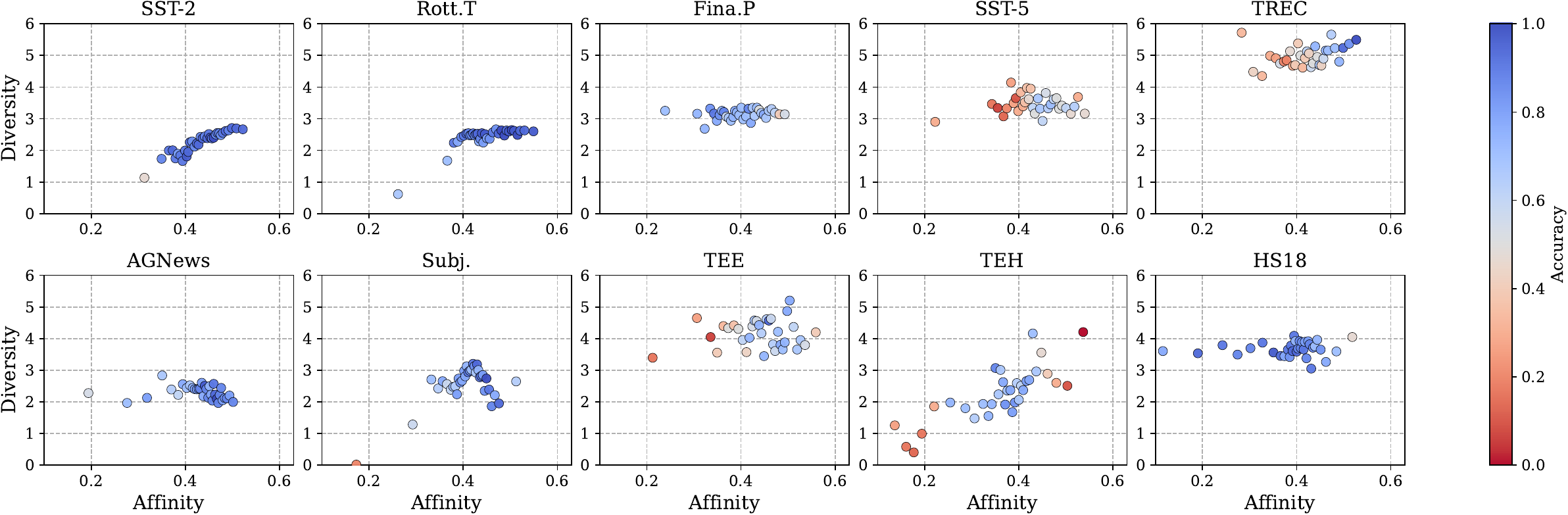}
    \caption{The Affinity and Diversity of the demonstrations. Colors refer to the accuracy of all classification tasks on $k=16$.}
    \label{fig:apd_1}
\end{figure*}

\begin{figure*}
    \centering
    \includegraphics[width=1\linewidth]{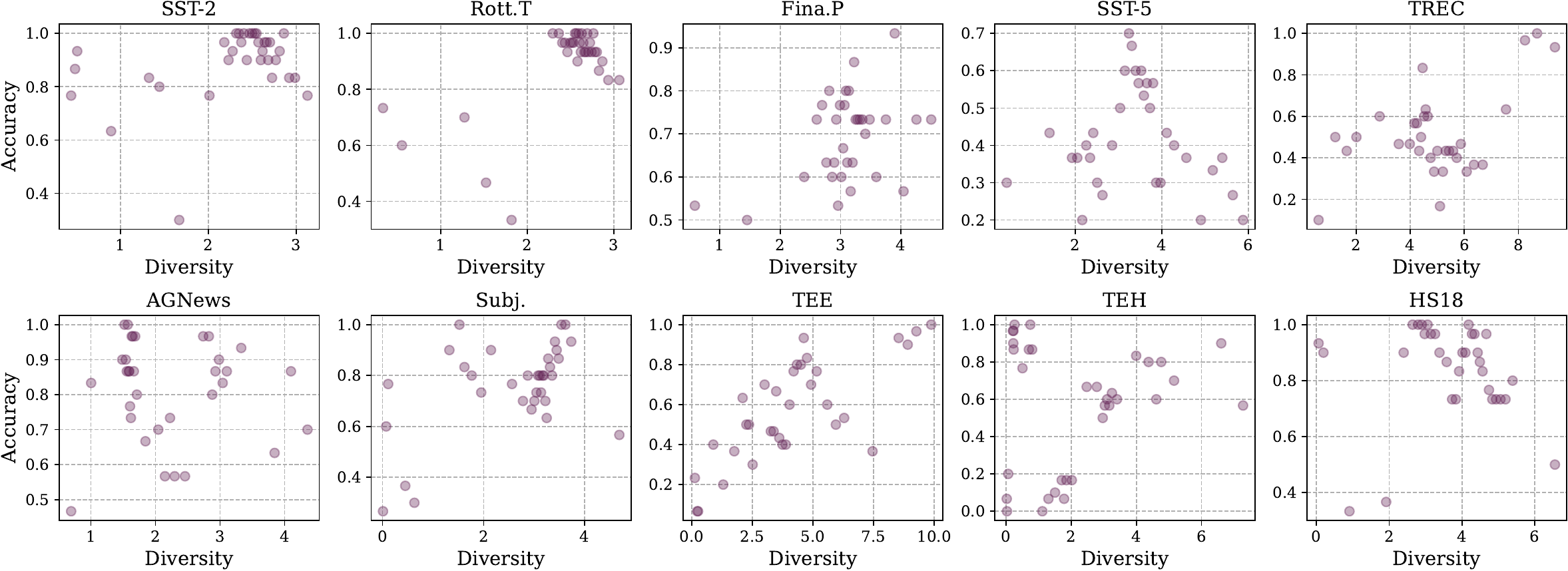}
    \caption{The tendency of diversity to accuracy on $k=16$. }
    \label{fig:apd_4}
\end{figure*}

\begin{figure*}
    \centering
    \includegraphics[width=1\linewidth]{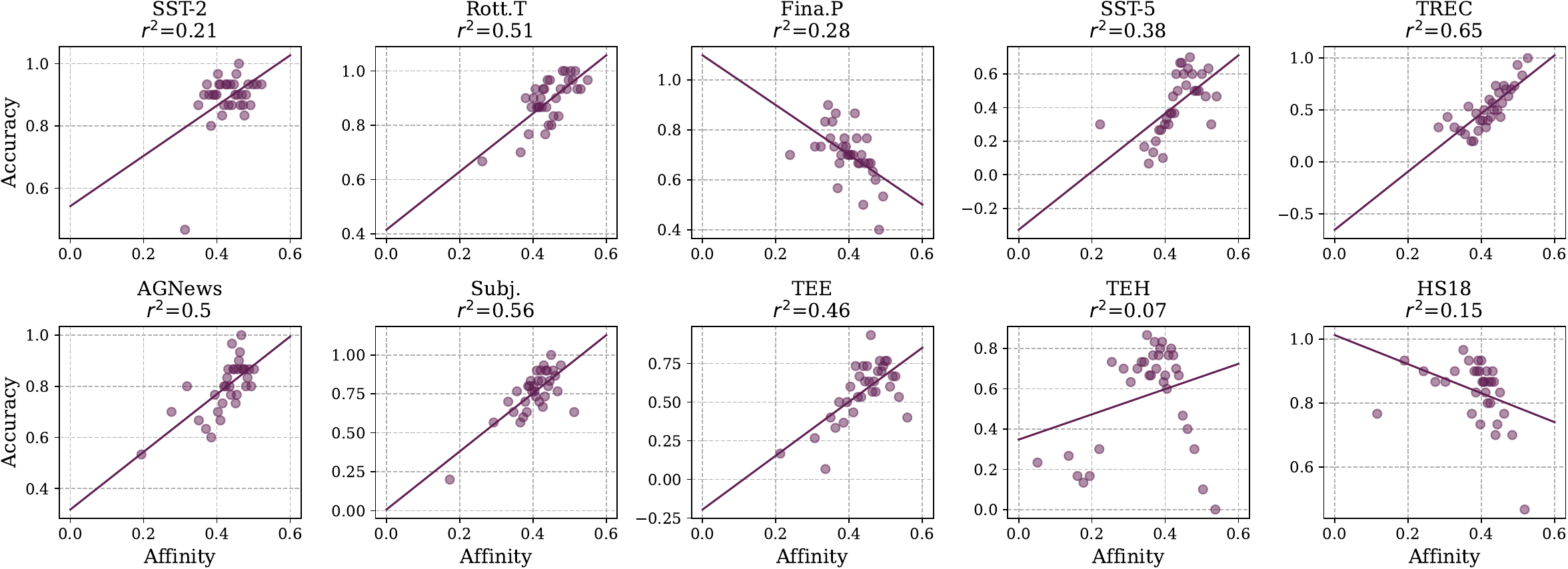}
    \caption{The tendency of affinity to accuracy on $k=16$.}
    \label{fig:apd_5}
\end{figure*}

\begin{figure*}
    \centering
    \includegraphics[width=1\linewidth]{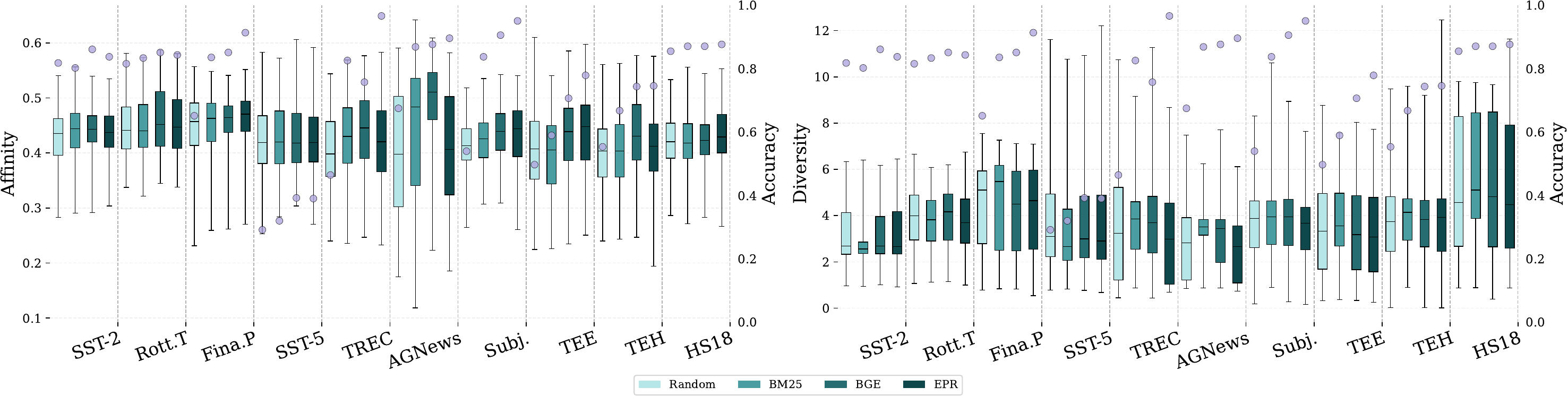}
    \caption{\textbf{Right}: The affinity of selected demonstrations by each selection method on $k=2$. \textbf{Right}: The diversity of selected demonstrations by each selection method on $k=2$.}
    \label{fig:apd_7}
\end{figure*}

\begin{figure*}
    \centering
    \includegraphics[width=1\linewidth]{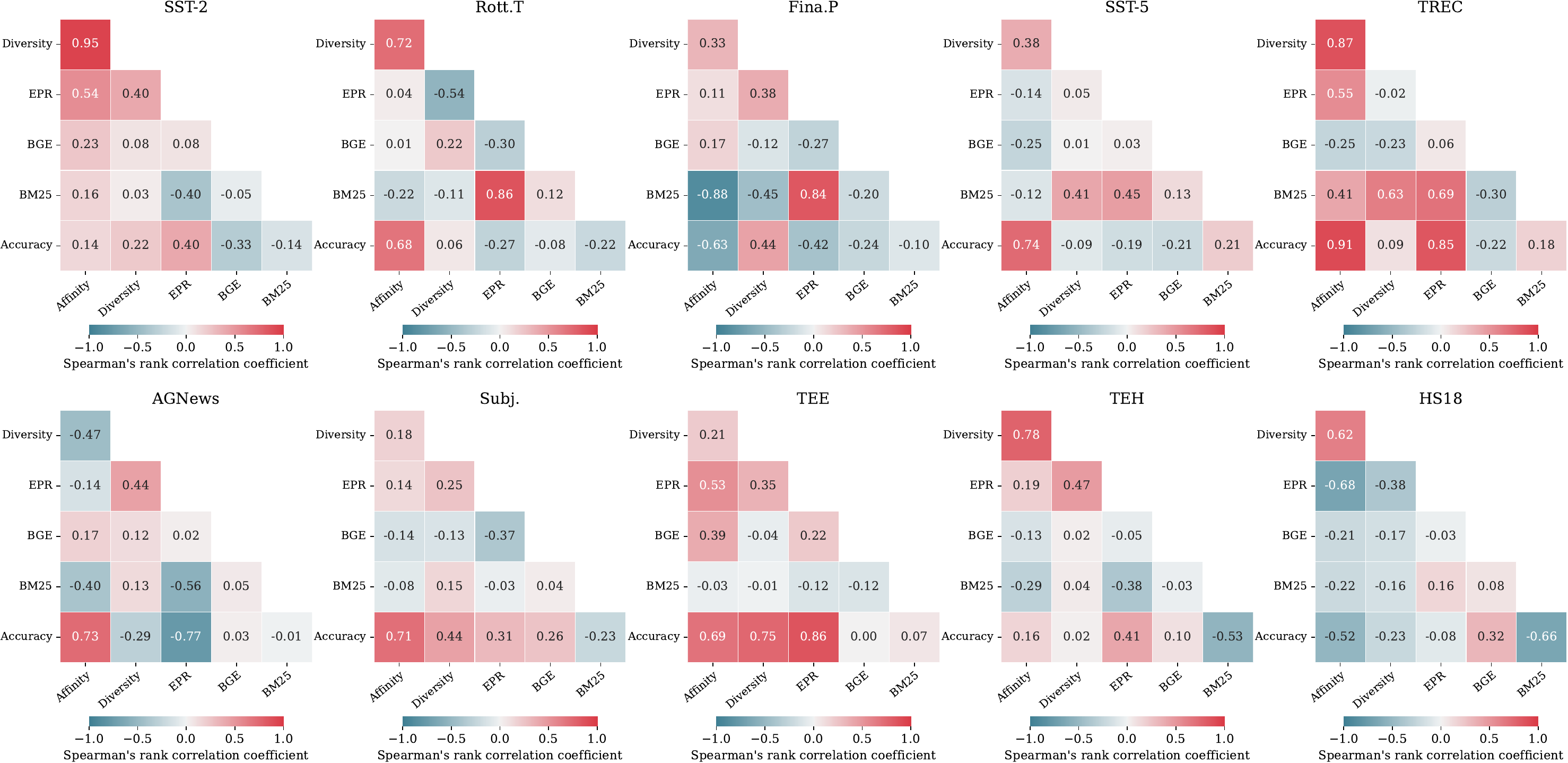}
    \caption{The Spearman's rank correlation coefficient of the similarity scores, affinity, diversity, and accuracy of $k=16$}
    \label{fig:apd_8}
\end{figure*}

The results of most experiments in the main text on other datasets are shown in Fig.~\ref{fig:apd_1},~\ref{fig:apd_4},~\ref{fig:apd_5},~\ref{fig:apd_7},~\ref{fig:apd_8}.

\section{Statements}
\subsection{License for Artifacts}
\paragraph{Models}
Llama 3 8B is under its specific license.
\paragraph{Datasets}
We list the open-source license for the datasets used in this paper as follows:
\begin{itemize}
    \item CC-by4.0: Tweet eval emotions, Tweet eval hate
    \item CC-BY-NC-SA-3.0: Financial phrasebank
    \item CC-BY-SA-3.0: Hate speech 18
    \item BSD: TREC, Subjective
    \item Unknown: GLUE-SST2, Rotten tomatoes, Stanford sentiment treebank, AGNews
\end{itemize}
\subsection{Statistics For Data}
We list the number of examples of datasets used in this paper as follows Table ~\ref{tab:dataset-size}.

\begin{table*}[t]
\caption{Raw dataset split size for each sub-dataset.}
\label{tab:dataset-size}
\resizebox{\textwidth}{!}{
\begin{tabular}{rllllllllll}
\toprule
 & SST2 & Rott.T & Fina.P & SST5 & TREC & AGNews & Subj. & TEE & TEH & HS18 \\ \midrule
Demonstration Set & 4096 & 4096 & 512 & 4096 & 4096 & 4096 & 4096 & 4096 & 3192 & 4096 \\
Test Set & 512 & 512 & 512 & 512 & 512 & 512 & 512 & 512 & 512 & 512 \\ \bottomrule
\end{tabular}}
\end{table*}

\subsection{AI Agent Usage}
AI Agents are only used for writing improving and grammar checking in this paper.

\end{document}